\pdfoutput=1

\documentclass[11pt]{article}

\usepackage[preprint]{acl}

\usepackage{times}
\usepackage{latexsym}
\usepackage{amsmath}
\usepackage{etex}
\usepackage{enumitem}
\usepackage[T1]{fontenc}
\usepackage{booktabs}
\usepackage{subcaption}
\usepackage[utf8]{inputenc}
\usepackage{authblk}

\usepackage{microtype}
\usepackage{tcolorbox}

\usepackage{inconsolata}
\usepackage{bm}
\usepackage{graphicx}

\newcommand{\shortname}{\textsc{M-SyMoN}}
\newcommand{\boldlongname}{\underline{M}ultilingual \underline{Sy}nopses of \underline{Mo}vie \underline{N}arratives}
\newcommand{\yidan}[1]{\textcolor{black}{#1}}
\newcommand{\jfcomment}[1]{\textcolor{black}{#1}}

\title{Multilingual Synopses of Movie Narratives: A Dataset for Vision-Language Story Understanding}

\author[1]{Yidan Sun}
\author[2]{Jianfei Yu}
\author[1]{Boyang Li}

\affil[1]{College of Computing and Data Science, Nanyang Technological University}
\affil[2]{School of Computer Science and Engineering, Nanjing University of Science and Technology}

\affil[ ]{\texttt{SUNY0053@e.ntu.edu.sg},  \texttt{jfyu@njust.edu.cn}, \texttt{boyang.li@ntu.edu.sg}}

\begin{document}
\maketitle

\begin{abstract}

Story video-text alignment, a core task in computational story understanding, aims to align video clips with corresponding sentences in their descriptions.
However, progress on the task has been held back by the scarcity of manually annotated video-text correspondence and the heavy concentration on English narrations of Hollywood movies.
To address these issues, in this paper, we \jfcomment{construct} a large-scale multilingual video story dataset named \boldlongname{} (\shortname{}), containing 13,166 movie summary videos from 7 languages, as well as manual annotation of fine-grained video-text correspondences for 101.5 hours of video. Training on the human annotated data from \shortname{} outperforms the SOTA methods by 15.7 and 16.2 percentage points on Clip Accuracy and Sentence IoU scores, respectively, demonstrating the effectiveness of the annotations. 
As benchmarks for future research, we create 6 baseline approaches with different multilingual training strategies, compare their performance in both intra-lingual and cross-lingual setups, exemplifying the challenges of multilingual video-text alignment. The dataset is released at: \url{https://github.com/insundaycathy/M-SyMoN}

\end{abstract}

\section{Introduction}
Computational story understanding aims to empower AI systems with the ability to delve into the intricacies of diverse stories, unlocking their deep semantics 
such as character motivations and intentions~\cite{emelin2020moral,rashkin-etal-2018-event2mind}, event structures~\cite{chambers-jurafsky-2008-even-chains,du2021learning}, \jfcomment{and} social relationships among story characters~\cite{elson2010extracting,chaturvedi2015modeling,kim2019frowning}. 
In recent years, computation story understanding has garnered significant research interest \cite{dong2023corrpus,2022tvshowguess,andrus2022enhanced,xu2022fantastic,han2023autoad} and many story understanding tasks \cite{wu2021towards,choi2021dramaqa,gu2023tevis} have emerged.

\begin{figure*}[t]
 \centering
 \includegraphics[width=\linewidth]{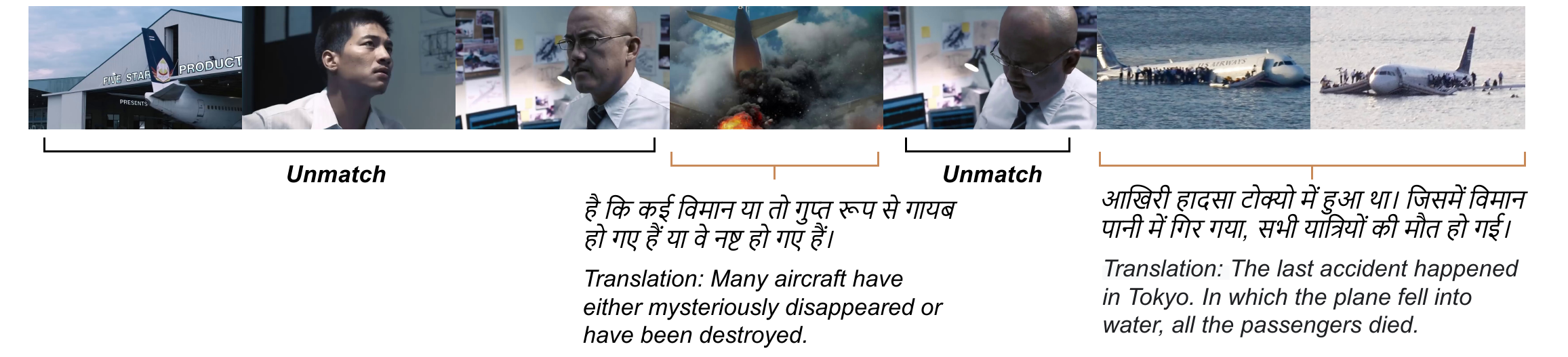}
  \caption{An example alignment between a video clip sequence and a sentence sequence from \shortname{}. One text chuck may correspond to several video clips, while some video clips may not match any textual description. The example is from  \url{https://youtu.be/n5v9hzSYxPQ}, a summary of the movie \emph{407 Dark Flight} (2012)}.
 \label{fig:dataset_exmaple}
 \end{figure*}

Story video-text alignment is a \jfcomment{fundamental} task of computational story understanding, which aims to find the best correspondence between a sequence of video clips and a sequence of sentences (see an alignment example in Figure~\ref{fig:dataset_exmaple}).
\jfcomment{Different from traditional video-text retrieval that relies on keyword or temporal cue matching~\cite{wang2021t2vlad}, story video-text alignment requires various story understanding abilities such as causal chain reasoning, mental state description, and long-range context understanding~\cite{sun2022synopses}.}
\jfcomment{Establishing such video-text correspondence will facilitate many applications such as text-to-video generation~\cite{liu2019cross,balaji2019conditional}, and visual story generation~\cite{huang2019hierarchically}.
Thus, the story video-text alignment task has recently attracted increasing attention~\cite{dogan2018neural,wang2021data,sun2023event}.
}

 \jfcomment{However, a major obstacle of this task stems from the scarcity of annotated data, as it is costly and time-consuming to manually annotate sentence-level alignments between text and clip sequences.
 The story understanding datasets used in previous studies are limited for several reasons.
 First, although the Large Scale Movie Description Challenge (LSMDC) dataset~\cite{rohrbach2017movie} manually aligns 158 hours of movies and audio descriptions, it only considers one-to-one matching between movie clips and audio descriptions and its audio descriptions are designed for visually impaired audiences, containing excessive details.
 Second, the YouTube Movie Summary (YMS) dataset~\cite{dogan2018neural} is small and only contains 6.7 hours of video and texts, and thus is typically employed for video-text alignment evaluation.
 Lastly, Hollywood movies and English narrations are dominant in 
 existing story understanding datasets~\cite{huang2020movienet,soldan2022mad,sun2022synopses,lu2023show}, neglecting the importance of language and cultural diversity.}

\jfcomment{To address the aforementioned limitations,} we construct a multilingual video story dataset named \boldlongname{} (\shortname{}). \shortname{} is sourced from movie recap videos on YouTube and contains 13,166 videos spanning 7 languages and totalling 2,136 hours. 
Furthermore, we \jfcomment{manually} annotated a portion of this dataset, providing exact video-text correspondence for 480 videos or 101.5 video hours.
\jfcomment{Compared to LSMDC and YMS, our annotated video-text alignment subset is large-scale, contains one-to-many matching and unmatched text or video clips, and covers 7 different languages.}


We further investigate the multi-lingual characteristics of the dataset, and make the following observations. 
First, in the intra-lingual setup, compared to translating all languages to English for training and inference, additional language-specific finetuning on weakly supervised data brings an average improvement of 5.4 percentage points across 7 languages.
Adding a small portion of the manual annotations further boosts performance. 
Second, in the cross-lingual setup, for source and target languages that are linguistically similar (e.g., Spanish to Portuguese or French), the transfer performance is generally good; for source and target languages that are different (e.g., Chinese to English or Hindi), the transfer performance is quite limited.
Third, 
an out-of-domain evaluation on the YMS dataset shows that training on the weakly supervised data from \shortname{} outperforms the state of the art methods by 12.3 and 13.2 percentage points on Clip Accuracy and Sentence IoU scores. Moreover, adding manually annotated video-text alignment data further improves the performance by 2.4 and 3.0 percentage points, indicating the utility of our annotated alignment data.

The main contributions of this work are summarized as follows:
\begin{itemize}
\setlength\itemsep{-0.2em}
    \item We \jfcomment{construct a large-scale} multilingual video story understanding dataset named \shortname{}, containing 13,166 videos in 7 languages and totaling 2,136 hours.
    \item We manually annotate the fine-grained alignment between video clips and sentences of 480 videos for a total of 101.5 hours.\footnote{The video URLs and video-text alignment annotations in \shortname{} will be made publicly available.} 
    \item We create \jfcomment{a number of multilingual video-text alignment methods} to benchmark the \shortname{} dataset. 
    \jfcomment{Extensive results on both \shortname{} and YMS demonstrate the significance of our multilingual dataset and the utility of the human video-text alignment annotations.}
\end{itemize}

\section{Related Work}
\subsection{Story Video-Text Alignment}

Story video-text alignment involves aligning sequences of video clips, typically from movies, with text captions \cite{cour2008movie,tapaswi2015book2movie,wang2021data,dogan2018neural}. The common approach is to first learn a video-text similarity metric and then calculate the alignment using dynamic programming (e.g., DTW) \cite{zhang2023aligning,dvornik2021drop}. Recent methods encode videos with 3D-convolution \cite{sun2023event} or ViT \cite{li2023strong} and texts with Transformer models \cite{han2022temporal,li2023strong,zhang2023exploring}, then optimize alignment via soft-DTW \cite{zhang2023exploring,han2022temporal} or calculate the alignment from video-text similarity \cite{zhang2023aligning}.

Due to the lack of sentence-level video-text alignment annotations, most models are trained with timestamp-based weak supervision. 
Although the LSMDC dataset~\cite{rohrbach2017movie} contains a large amount of manual annotation, its overly fine-grained text data interferes with event understanding \cite{sun2022synopses}; it also lacks one-to-many matching and unmatched items. 
The YMS dataset~\cite{dogan2018neural} contains only 6.7 hours of movie summaries with exact video-text alignment and cannot serve as a comprehensive test benchmark.

\subsection{Movie Story Datasets}
Movies are a popular source of video-text story content. LSMDC \cite{rohrbach2017movie} and Movie Audio Description (MAD) \cite{soldan2022mad} provide movie clips with audio descriptions for the visually impaired.
Although the audio descriptions are accurate, they deviate significantly from realistic styles of story narration. 
The Condensed Movie Descriptions (CMD) dataset \cite{bain2020condensed} offers 7 to 11 key clips per movie with one-sentence descriptions. 
The Pororo dataset~\cite{kim2017deepstory} includes 20-minute cartoon episodes with in-show conversations and human-written descriptions. 
Although CMD and Pororo captions match the clips, they may not form complete storylines. The CVSV \cite{lu2023show}, YMS \cite{dogan2018neural}, and SyMoN \cite{sun2022synopses} datasets collect YouTube movie summaries, similar to \shortname{}, but CVSV and SyMoN lack human annotations.

\jfcomment{Although there are many story understanding dataset, multilingual video story datasets are scarce. To our knowledge, the only dataset is Movie101v2 \cite{yue2024movie101v2},} which contains 46K Chinese video-caption pairs from 203 movies, where the Chinese text are translated to English using GPT-3.5-turbo. The dataset also lacks human-annotated video-text correspondence. 
In comparison,  \shortname{} is large-scale and multilingual, containing movie summary videos in 7 languages and 101.5 hours of video with exact video-text correspondences. 

\subsection{Multilingual Story Understanding}




\jfcomment{There are several tasks related to multilingual story understanding, including} event-causal inference \cite{lai2022meci}, story question answering \cite{ateeq2023arabic}, story tag classification \cite{tikhonov2021storydb}, and story generation \cite{razumovskaia2022little}.
\jfcomment{However, most of them merely focus on text rather than video stories.}

Methods for multilingual story understanding mainly fall into two categories: (1) translating training/test data to English \cite{ponti2020xcopa} or prompts to the target language \cite{lin2022few}, and (2) directly finetune on non-English text using a multilingual Pretrained Language Model (PLM) \cite{tikhonov2021storydb} or a PLM for the target language \cite{lai2022meci}. In this paper, we benchmark \shortname{} on multilingual story video-text alignment with both types of methods.

\section{Task Formulation}
Given a consecutive sequence of video clips $V = (v_1,v_2,...,v_m)$ and a consecutive sequence of sentences $T = (t_1,t_2,..t_n)$, 
\jfcomment{the video-text alignment task aims to find the alignment between these two sequences by learning a function $f$ that maps each input sentence $t_i$ to its corresponding video clips:
\begin{equation}
\mathcal{P}=\{ \dots, (t_i, v_{f(i)}), \dots \}
\end{equation}
where $f(i)$ refers to the indexes of the video clip aligned with the sentence $t_i$. Note that $f(i)$ can be \textit{None}, a single index, or an index sequence, denoting that there is no video clip, one video clip, or multiple video clips aligned with $t_i$, respectively.}

In the task formulation, we make the simplifying assumption that multiple video clips can be aligned with one sentence, but a clip cannot be aligned with more than one sentence. The reason is that a video clip is typically very short (around 2.4 seconds) and clips outnumber sentences, so it is rare that one clip aligns with more than one sentence. 



\begin{table*}[!tp]
\centering
\small
\begin{tabular}{@{}p{5.5cm}p{1.0cm}p{1.0cm}p{1.0cm}p{1.0cm}p{1.2cm}p{1.0cm}p{1.0cm}@{}}
\toprule
Languages & English     &  Chinese & Spanish & French & Portuguese & Hindi & Russian\\
\midrule
Video Count & 5,193  & 2,683 & 1,595 & 1,193 & 1,070 & 749 & 683\\
Video Hours & 869 & 390 & 285 & 102 & 209 & 173 & 108\\
Vocabulary Size & 40,116 & 6,269 &50,050&33,967&40,676 & 20,896 & 64,827\\
Average Video Length (minutes) & 9.5 & 8.7 & 10.7 & 5.1 & 11.7 & 13.9 & 9.5 \\
Average number of Sentences in a Narration & 131 & 78.8 & 82.6 & 138 & 134 & 126 & 151 \\
Average number of Words in a Narration & 1,717 & 3,234 & 1,923 & 2,226 & 2,737 & 2,656 & 1,623\\
Movie Count & 2,440 & 1,236 & 1,217& 877 & 811 & 462 & 542\\
Number of Annotated Videos & 98 & 30 & 75 & 84 & 57 & 63 & 72 \\
Annotated Video Hours & 23.42 & 7.23 & 14.17 & 14.17 & 14.17 & 14.17 & 14.17 \\

\bottomrule
\end{tabular}
\caption{The statistics of \shortname{}. For Chinese, we regard the number of unique characters as its vocabulary size.}
\label{tab:dataset_statistics}
\end{table*}

\section{Dataset Construction}
In this section, we describe \jfcomment{the details of constructing the \shortname{} dataset.}

\subsection{Data Source and Statistics}

For data collection, we first identify YouTube channels with movie recap videos in the target languages. Keywords such as ``movie summary in <language>'' and ``movie recap in <language>'' are used to search for the channels. 
We then download all videos and their accompanying subtitles from the identified channels. 
Videos without subtitles and that are not movie summaries are discarded. 

This yields 13,166 videos or 2,136 video hours in 7 languages, including English, Chinese, Spanish, French, Portuguese, Hindi, and Russian. 
We list the statistics in Table \ref{tab:dataset_statistics}. 
\shortname{} contains summaries for 5,960 movies and TV shows, of which 1,515 have more than one summary. 

\subsection{Human Annotation}
We hire a professional team of annotators \jfcomment{with rich} translation experience from Flitto\footnote{https://www.flitto.com} to annotate exact video-text correspondence in \shortname{}. In total, we annotated 480 videos spanning 101.5 hours. The amount of annotated videos in each language are in the last two lines of Table \ref{tab:dataset_statistics}.

Before human annotation, we automatically divided the video subtitle text into sentences. For each sentence, we ask the annotators to locate the start and end times of the video segment described by the sentence. 
If a sentence is not grounded in the video, it is marked as ``unmatched''. 
We removed the audio from the videos to eliminate shortcut features for alignment from the audio. 
Note that human-annotated video-text correspondence is different from temporal correspondence, because the narration in narrated movie summary videos is sometimes faster or slower than the video, and some narrator may add commentary not grounded in the video.

\noindent \textbf{Annotation Quality.}
To evaluate annotation quality, 
\jfcomment{we also employ a student who is familiar with the task to} annotate a small validation set for each language. 
The annotation quality is evaluated as the IoU between the durations annotated by the student and the durations annotated by the annotators. If the IoU for a particular language falls below 60\%, the annotators were asked to redo the annotations until the IoU reaches 60\%. The final average IoU across 7 languages is 83.1\%, indicating substantial agreement between the annotators. See Appendix \ref{app:annotation} for annotation details.

\subsection{Data Preprocessing}
\textbf{Multilingual Punctuation Restoration.}
We acquire text descriptions directly from YouTube subtitles. 
In some cases, the text \jfcomment{descriptions are derived from automatic speech recognition tools} and are unpunctuated. As punctuated texts are required for downstream tasks, we train a Transformer-based model to restore punctuations in all unpunctuated text. We do not use off-the-shelf punctuation restoration models because most models do not include all the languages in \shortname{} \cite{chordia2021punktuator,frank2021fullstop}. 
See Appendix~\ref{app:punc_restore} for details on the punctuation restoration model. 


\noindent \textbf{Scene Segmentation.}
We divide the videos into clips using TransNet V2~\cite{souvcek2020transnet} which detects hard camera cuts. Each clip, containing a continuous shot between two hard camera cuts, is roughly 2.4 seconds long.

\noindent \textbf{Offensive Language Filtering.}
To the best of our abilities, we created a list of offensive terms and filtered out videos containing those phrases.   

\subsection{Dataset Split}
\label{sec:data_split}
We divide \shortname{} into two parts: (1) the human annotated portion: 480 videos from 7 languages manually annotated with exact video-text correspondence; (2) the weakly supervised portion: The entire \shortname{} dataset after removing all movies that appear in the annotated portion (see Appendix \ref{app:remove_overlap} for details). 

For the weakly supervised data, each video clip is matched to the text segment that spans its duration, producing a rough video-text correspondence regarded as the weakly supervised training set. 

For the human annotated data, we further split it into a training set (20\%), a validation set (20\%), and a test set (60\%), with the portions distributed evenly among 7 languages. Table \ref{tab:dataset_split} shows the number of clip-sentence pairs in each split.

\begin{table*}[!tp]
\centering
\small
\begin{tabular}{llllllll}
\toprule
Languages & English     &  Chinese & Spanish & French & Portuguese & Hindi & Russian\\
\midrule

Weakly supervised training set & 730,112 & 365,312& 166,530 & 60,676 & 127,788 & 90,560 & 84,556\\
Supervised training set &6,649 &1,860 &5,989 &5,799 & 4,205& 7,152 &5,789\\
Validation set & 7,201 &1,616 & 4,066 &4,717 & 6,865 & 7,603 & 8,474 \\
Test set & 16,208 & 5,441  & 15,185 & 12,560 & 15,356 & 19,652 & 14,664\\ 

\bottomrule
\end{tabular}
\caption{The number of clip-sentence pairs in each data split.}
\label{tab:dataset_split}
\end{table*}

\subsection{Dataset Analysis}
\label{sec:dataset-analysis}
We hypothesize that the choices of movies in each language may have cultural characteristics. 
Due to space limitations, we analyze the country of release and themes of the movies in this section. Additional statistics of the genre, year of release, language of release, and narrative structures are illustrated in Appendix \ref{app:analysis}.

As most movie metadata are not available from the movie recap videos, we use ChatGPT (gpt-3.5-turbo-0125) to acquire the information. Specifically, we input the video title, the video-level description featured on YouTube, and first 5 sentences of the movie narration into ChatGPT and ask ChatGPT to output the movie title and metadata. 

\begin{figure}[!tp]
    \centering
    \includegraphics[width=0.5\textwidth]{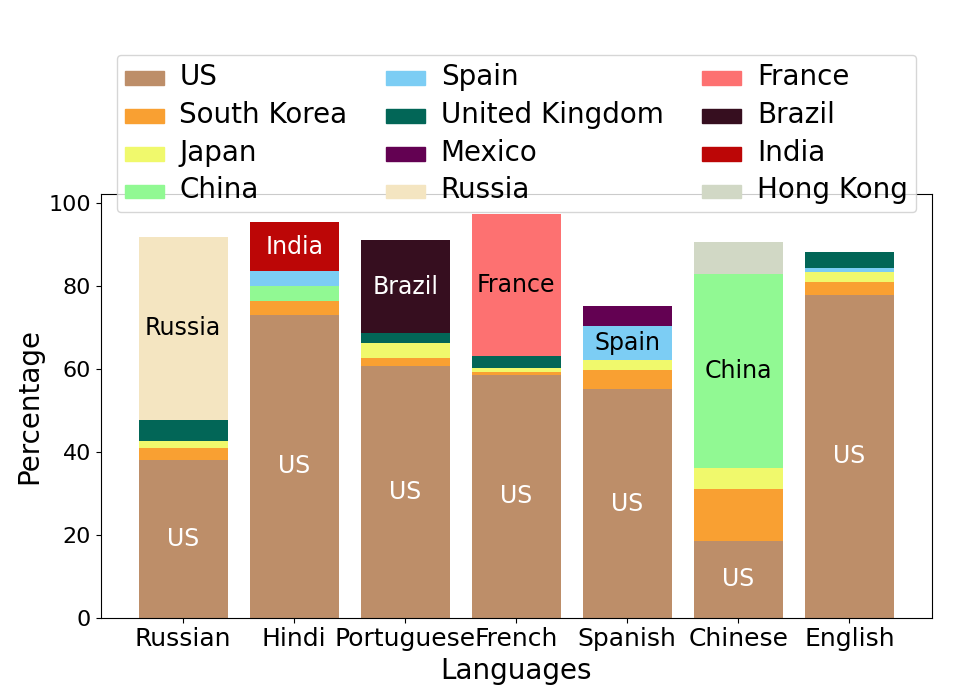}
    \caption{The top 5 countries of release for movies recapped in each language.}
    \label{fig:country}
\end{figure}

\jfcomment{Figure \ref{fig:country} reveals interesting correlation between the language and the origin of the movie. The movie summaries in each language are typically about domestic movies (i.e., movies from countries where the language is widely used) or from the United States, reflecting the influence of Hollywood. For instance, 22.5\% of Portuguese videos are about Brazilian movies.  
Interestingly, large percentage of videos in Chinese (46.9\% from China Mainland, 7.6\% from Hong Kong) and Russian (44.0\%) describe domestic movies than other languages. In comparison, on average 20.4\% of videos in other non-english languages describe domestic movies. }

\begin{figure}[!tp]
    \centering
    \includegraphics[width=0.47\textwidth]{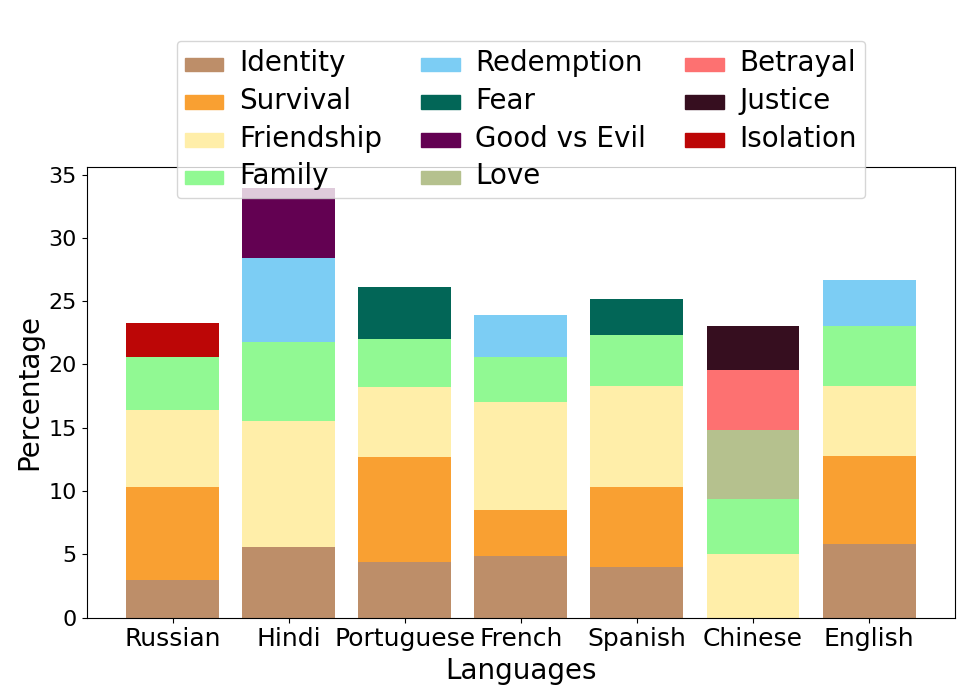}
    \caption{The top 5 themes for movies recapped in each language.}
    \label{fig:theme}
\end{figure}

Figure \ref{fig:theme} reveals variations in thematic preferences across languages. While themes like ``Friendship'' and ``Family'' appear universally popular, others exhibit uneven distributions. For instance, 2.7\% (vs. 0.7\% in other languages on average) of Russian movies have the theme  "Isolation". The theme ``Identity'' accounts for only 1.8\% Chinese videos (vs. an average of 4.6\% in all other languages). These differences highlight the importance of representation of multiple cultures and languages in story understanding datasets. 

\section{Methodology}
\jfcomment{This section first details our base model for video-text alignment and then introduces its several variants with different multilingual training strategies.}

\subsection{Base Model}
\jfcomment{To achieve the sentence-level alignments between text and clip sequences, we adopt a three-stage framework widely used in recent studies~\cite{dvornik2021drop,zhang2023aligning}.
Specifically, our model first employs video and text encoders to obtain the video clip representation and the text representation as follows.
}

\noindent \textbf{Video Encoder}.
\jfcomment{Given a video with $M$ clips, we represent each clip by randomly selecting three frames from it and dividing each frame into $H \times W$ patches. 
We then utilize a trainable MLP projection layer to map these patches into video tokens, which are fed into the video encoder Swin Transformer~\cite{liu2021swin} to obtain the visual representation of each frame.
An average pooling layer is applied over the three frames of each video clip to obtain the video clip representation.
}

\noindent \textbf{Text Encoder}.
\jfcomment{
Because our goal is to perform multilingual video-text alignment in this paper, we adopt a pre-trained cross-lingual cross-modal model named CCLM~\cite{zeng2022cross} to initialize the parameters in Swin Transformer and the text encoder.
Following CCLM, we employ a pre-trained multi-layer Transformer, i.e., XLM-R~\cite{conneau2019unsupervised} as the text encoder.
We then regard the last hidden representation of the \textit{[CLS]} token as the representation of each sentence.
Formally, let $\bm t_{i}$ and $\bm v_{i}$ denote the encoded features for the $i^{\text{th}}$ sentence and the $i^{\text{th}}$ video clip.
}

For every clip-text pair, we randomly sample $K$ hard negatives from the same video. With the sampled negative text features and video features, we finetune the encoders by minimizing the contrastive InfoNCELoss
\citep{oord2018representation} below:
\begin{equation}
\begin{aligned}
\mathcal{L}_{\text{InfoNCE}} = &\frac{1}{N} \sum_{i=1}^{N} \left[ - \log \left( \frac{\exp(\bm{v}_{i}^\top \bm{t}_{i} / \tau)}{\sum_{j=1}^{K} \exp(\bm{v}_{i}^\top \bm{t}_{j} / \tau)} \right) \right.\\
&\left. - \log \left( \frac{\exp(\bm{v}_{i}^\top \bm{t}_{i} / \tau)}{\sum_{j=1}^{K} \exp(\bm{v}_{j}^\top \bm{t}_{i} / \tau)} \right) \right]
\end{aligned}
\end{equation}
where $N$ is the total number of training samples and $K$ is the number of negative samples.

\jfcomment{During inference, we first obtain the representation of each clip and sentence, and then calculate the cosine similarity between every clip-sentence pair.
Finally, we find the global clip-sentence alignment from the similarity scores of each pair by resorting to a sequence alignment algorithm named Double Drop Dynamic Time Warp (Drop-DTW) \cite{dvornik2021drop},  detailed in Appendix \ref{app:drop_dtw}.}

\begin{table*}[!tp]
\centering
\small
\setlength{\belowcaptionskip}{-0.4cm}
\begin{tabular}{@{}p{4.1cm}p{1.0cm}p{1.0cm}p{1.0cm}p{1.0cm}p{1.2cm}p{1.0cm}p{1.0cm}p{1.0cm}@{}}
\toprule
  & English    &  Chinese & Spanish & French & Portuguese & Hindi & Russian & Average\\
\midrule
\multicolumn{9}{c}{\textit{Weakly supervised}} \\
CCLM-multilingual  & 9.3 & 16.3 & 11.8 & 8.8 & 9.9 & 5.8 & 9.3 & 10.2\\
CCLM-individual & 26.6 & 29.3 & 17.4 & 16.6 & 16.3 & 10.1 &14.6 & 18.7 \\ 
CCLM-translate  & 22.2 & 20.1 & 16.0 & 17.5 & 14.1 & 11.2 & 13.8 & 16.4 \\
CCLM-two-stage  & \textit{\textbf{26.9}} & \textit{\textbf{37.7}} & \textit{\textbf{19.1}} & \textit{\textbf{20.2}} & \textit{\textbf{18.7}} & \textbf{13.1} & \textit{\textbf{17.0}} & \textit{\textbf{21.8}}\\
\midrule
\multicolumn{9}{c}{\textit{Supervised}} \\
CCLM-translate-supervision &24.3 & 20.1 & 17.8 & 19.7 & 16.0 & 12.6 & 15.1 & 17.9 \\ 
CCLM-two-stage-supervision & \textbf{27.7} & \textbf{38.9} &  \textbf{19.8} & \textbf{21.0} &\textbf{19.5} & \textit{\textbf{12.9}} & \textbf{17.5} & \textbf{22.5}\\
\bottomrule
\end{tabular}
\caption{Intra-lingual results based on F1 score. The ``Average'' is the mean of 7 languages. The highest number in each column is in \textbf{bold}, the second highest is in \textit{\textbf{bold italic}}.}
\label{tab:iid_results}
\end{table*}

\subsection{Multilingual Finetuning Methods}
\jfcomment{With} the base model, we benchmark the \shortname{} dataset by establishing 4 weakly supervised methods, which share the same model architecture \jfcomment{but are trained on the entire weakly supervised training set of \shortname{} in Section \ref{sec:data_split}} with different training strategies:
\begin{itemize}
    \item Multilingual training (CCLM-multilingual). We \jfcomment{combine the weakly supervised training sets of all 7 languages and train a unified multilingual model on the combined dataset.}
    \item Individual training (CCLM-individual). Using weakly supervised training data, we finetune the base model on each language \jfcomment{and obtain 7 language-specific models}.
    \item Translational training (CCLM-translate). We first translate all the weakly supervised training data and the test data in \shortname{} to English with off-the-shelf translation model NLLB-200 \cite{costa2022no}. We then finetune the base model on the translated data to obtain a \jfcomment{unified model}. 
    \item Two-stage training (CCLM-two-stage). On top of the trained CCLM-translate model, we further finetune on each language separately to obtain 7 language-specific models. 
\end{itemize}

\jfcomment{To show the usefulness of our manually annotated video-text alignment dataset, we further finetune two aforementioned methods on the human annotated training set in  Section \ref{sec:data_split}:}
\begin{itemize}
    \item 
    \jfcomment{CCLM-translate-supervision. We translate the non-English languages in the human annotated data to English and finetune the CCLM-translate model on the translated data.} 
    \item \jfcomment{CCLM-two-stage-supervision. We finetune the CCLM-two-stage model on the human annotated data of each language separately.}
\end{itemize}

\section{Experiments}
\jfcomment{In this section, we conduct extensive experiments in intra-lingual, cross-lingual, and out-of-domain setups to show the usefulness of \shortname{}.}

\subsection{Experimental Setup}

For CCLM-translate and CCLM-multilingual, we initialize their parameters from the pre-trained CCLM-base model~\cite{zeng2022cross} and finetune them for 20 epochs with an initial learning rate of $4 \times 10^{-5}$ and cosine learning rate decay. 
For CCLM-two-stage, we initialize its parameters from the CCLM-translate model and finetune it for 20 epochs with an initial learning rate of $4 \times 10^{-6}$ and cosine learning rate decay. 
Random augmentation and weight decay of 0.2 are applied. 

\subsection{Evaluation Metrics}
Following \citet{dogan2018neural}, we use two evaluation metrics. Clip accuracy is defined as the temporal proportion of correctly aligned video segments. 
Sentence IoU \yidan{\citep{jaccard1908nouvelles}} is defined as the intersection-over-union between the aligned video durations and the groundtruth durations. 
Due to space limitations, we mainly report the F1 score, i.e. the harmonic mean between Clip Accuracy and Sentence IoU in this section and defer detailed results to Appendix \ref{app:clip_acc}.




\subsection{Intra-Lingual Results}
\label{sec:iid_results}
\jfcomment{We report the intra-lingual results in Table \ref{tab:iid_results}. Here, each method is trained and evaluated on data in the same language.}

First, we find that training \jfcomment{the base model} on all languages together performs significantly worse than training on each language separately. Specifically, CCLM-individual outperforms CCLM-multilingual by 4.3-17.3 percentage points.
\jfcomment{This suggests that language-specific features are important for multilingual video-text alignment.}
Second, the CCLM-translate baseline is on par with CCLM-individual and significantly outperforms CCLM-multilingual. 
\jfcomment{Moreover, the two-stage baseline (i.e., CCLM-two-stage) leverages the benefits of pre-training on a large multilingual dataset and subsequent finetuning on a specific language dataset, which outperforms CCLM-individual and CCLM-translate by 3.1 and 5.4 percentage points on average.}
\jfcomment{Finally, after finetuning with our manually annotated data, it yields consistent improvements on top of both CCLM-translate and CCLM-two-stage models across the 7 languages, and CCLM-two-stage-supervision achieves the best average performance among all the compared methods.
These observations demonstrate the usefulness of the manually annotated video-text alignment data.}


\begin{figure}[!tp]
    \centering
    \includegraphics[width=0.45\textwidth]{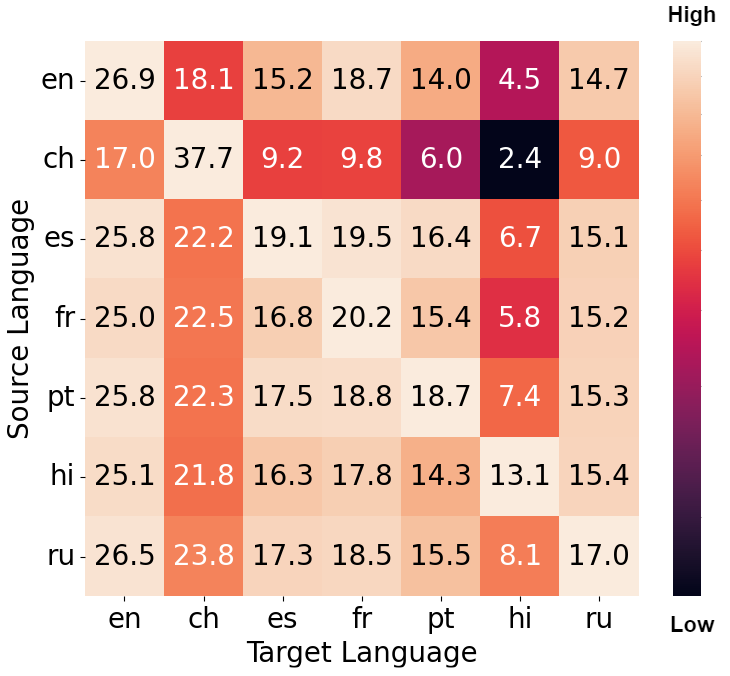}
    \caption{Cross-lingual transfer results of CCLM-two-stage based on F1 score. The language names are abbreviated as: English =``en'', Chinese = ``ch'', Spanish = ``es'', French = ``fr'', Portuguese  ``pt'', Hindi = ``hi'', Russian = ``ru''.}
    \label{fig:cross-lingual}
\end{figure}

\subsection{Cross-Lingual Transfer Results}
\label{sec:cross-lingual}
\jfcomment{We employ the CCLM-two-stage model to evaluate the cross-lingual performance of every source-target language pair, }see Figure \ref{fig:cross-lingual} for results. 

First, the highest value of each column appears on the diagonal.
This is intuitive because the model is trained and evaluated on the same language. 
Second, linguistically similar languages transfer well to each other. 
For example, related languages like Spanish and Portuguese, 
generally obtain good cross-lingual performance.
This likely stems from shared vocabulary, sub-word tokens, and grammatical structures.
Furthermore, Chinese does not transfer well to any other language, possibly due to differences in linguistic construction and movie distributions (See \S \ref{sec:dataset-analysis}) . 
\jfcomment{Lastly, we observe} it is difficult to transfer from any language to Hindi, possibly because the CCLM pre-training data does not contain Hindi, which hurts representation learning. 
\begin{table*}[!tp]
\centering
\small
\setlength{\belowcaptionskip}{-0.1cm}
\begin{tabular}{@{}p{4.1cm}p{1.0cm}p{1.0cm}p{1.0cm}p{1.0cm}p{1.2cm}p{1.0cm}p{1.0cm}p{1.0cm}@{}}
\toprule
  & English    &  Chinese & Spanish & French & Portuguese & Hindi & Russian & Average\\
\midrule
\multicolumn{9}{c}{\textit{Without Timestamps of Text}} \\
CCLM-two-stage-supervision & 27.7 & 38.9 &  19.8 & 21.0 &19.5 & 12.9 & 17.5 & 22.5\\
\midrule
\multicolumn{9}{c}{\textit{With Timestamps of Text}} \\
Transcript baseline  &  34.4 &  42.6 & 19.1 &  26.2 & \textbf{40.6} &  20.1 & \textbf{27.3} & 30.1\\
CCLM-2S-supervision-time & \textbf{34.6} & \textbf{46.2} & \textbf{28.5} & \textbf{27.7} & 39.8 & \textbf{25.4} & 26.4 & \textbf{32.7}\\
\bottomrule
\end{tabular}
\caption{Intra-lingual video-text alignment results with transcript temporal information.}
\label{tab:temporal_results}
\end{table*}

\subsection{Effects of Timestamps of Narration Text}
\label{sec:temporal_info}

All previous models do not utilize the timestamps of the narration texts as input features or during inference. As such timestamps are only available after a video summary is made, models relying on timestamps as input features cannot handle general alignment tasks, such as aligning plot summaries with movies. 

In this section, we explore the effectiveness of these timestamps during inference. Specifically, we use the CCLM-two-stage-supervised model to calculate video-text similarity. During DTW, we constrain video clips to match only sentences within a 1-second window of their timestamps. This model is named CCLM-2S-supervised-time. We also create a transcript baseline that relies solely on the timestamps; we simply align video clips to the sentence whose timestamp falls in their duration. 

Table \ref{tab:temporal_results} summarizes the results. First,
the transcript baseline achieves an average F1 score of only 30.1, indicating significant discrepancies between the temporal transcripts and our annotations. This demonstrates the noisy nature of YouTube transcripts and underscores the importance of precise annotations for accurate evaluation. Second, CCLM-2S-supervised-time outperforms CCLM-two-stage-supervised by 10.2 percentage points, showing that transcript temporal information improves alignment. Finally, CCLM-2S-supervised-time surpasses the transcript baseline by 2.6 percentage points, demonstrating that our model learns video-text correspondences beyond the transcripts.

\subsection{Transfer to YMS}
\label{sec:yms}
In this section, we extend our analysis to investigate if models trained on \shortname{} can generalize to other video-text alignment benchmarks, such as the out-of-domain English benchmark dataset YMS~\cite{dogan2018neural}.

As shown in Table \ref{tab:yms}, the large scale \shortname{} dataset significantly improves the performance of YMS. 
\jfcomment{First, we can observe that finetuning the base model on the human-annotated training set of YMS (i.e., CCLM-individual (YMS)) is already on par with the state-of-the-art performance, demonstrating the effectiveness of our proposed base model.
Second, compared to CCLM-individual (YMS), only training on the weakly supervised portion of \shortname{} improves performance by 13.7 and 12.7 percentage points on Clip Accuracy and Sentence IoU, respectively.
This suggests the value of the large-scale weakly-supervised data in \shortname{}.
Moreover, additional finetuning on the human-annotated data from \shortname{}, i.e., CCLM-two-stage-supervision (English), further boosts the performance by 2.4 and 3.0 percentage points on Clip Accuracy and Sentence IoU, respectively.
These observations highlight the significance and generalizability of our human-annotated video-text correspondence dataset, indicating the potential of \shortname{} for advancing research in video story understanding.}


\begin{table}[!tp]
\centering
\small
\begin{tabular}{@{}p{5.1cm}p{0.9cm}p{0.9cm}@{}}
\toprule
 & Clip Acc. & Sent IoU \\
\midrule
NeuMatch \cite{dogan2018neural} & 12.0  & 10.4  \\ 
\citet{wang2021data} & 30.6 & 12.8  \\ 
\citet{sun2023event} & 23.2 & 18.4  \\
\midrule
CCLM-individual (YMS) & 29.2 & 18.9 \\
\midrule
CCLM-multilingual& 13.3 & 6.2  \\
CCLM-individual (English) & 35.8 & 25.0  \\
CCLM-translate &25.2 & 16.7  \\
CCLM-two-stage (English) & 42.9 & 31.6\\
CCLM-two-stage-supervision (English) &\textbf{45.3} & \textbf{34.6}  \\

\bottomrule
\end{tabular}
\caption{Out-of-domain results on the YMS dataset. CCLM-individual (YMS) is the base model finetuned on the human-annotated training set of YMS.}
\label{tab:yms}
\end{table}


\subsection{Discussion}
The experiments in this paper demonstrate the value of \shortname{} as a multilingual and multimodal story understanding dataset. First, \shortname{} is the first dataset that offers large-scale human annotations for multilingual story video-text alignment, and such annotations deliver pratical benefits. As shown in \S \ref{sec:temporal_info}, the timestamps in the YouTube transcripts are noisy, and human annotations are necessary for accurate benchmarking. Additionally, the human-annotated training set provides valuable supervision signals. In \S \ref{sec:iid_results}, we achieve a notable 3.2\% relative improvement from human annotation data that amounts to only 2.2\% of the training data. Moreover, finetuning on supervised data from \shortname{} lead significant improvement on the out-of-domain YMS dataset in \S\ref{sec:yms}. 
Finally, \shortname{} provides fertile ground for linguistic and cultural studies of story elements. For example, cross-lingual transfer experiments (\S \ref{sec:cross-lingual}) lends support to existing linguistic theories on positive and negative transfer in language learning \cite{eronen2023zero,odlin1989language} as linguistically similar languages transfer well to each other. The analysis of movie themes for different languages (Figure \ref{fig:theme}) demonstrates the cultural relevance of \shortname{}.

\section{Conclusion}

\jfcomment{In this paper, we introduced \shortname{}, a large-scale multilingual video story dataset.
It contains 13,166 movie summary videos from 7 languages, featuring 480 videos with human-annotated video-text correspondence. 
We established several multilingual multimodal baselines to benchmark \shortname{}. Experimental results show the value of \shortname{} for video story understanding.
Its size, multilingual nature, and rich alignment annotations make \shortname{} a valuable contribution.}

\section*{Acknowledgments}
We gratefully acknowledge the support by the Nanyang Associate Professorship and the National Research Foundation Fellowship (NRF-NRFF13-2021-0006), Singapore. Any opinions, findings, conclusions, or
recommendations expressed in this material are
those of the authors and do not reflect the views of
the funding agencies.

\section*{Limitations}

The preprocessing of \shortname{} involves several automatic techniques that may introduce noise. First, some text descriptions are generated using Google automatic speech recognition. Automatic punctuation restoration is then applied to the text. For the translation baselines, the text is further translated into English. This pipeline may propagate errors. However, due to the size and complexity of \shortname{}, manual processing is impractical. We acknowledge the potential for improvement in preprocessing steps as better techniques become available.

Additionally, human annotation of video-text correspondence in story videos can be ambiguous. Annotators received a list of instructions for ambiguous cases. For example, when text describes a character's emotional state, annotators are instructed to match only when the emotion is evident from the character's expression or action. In cases outside the provided guidelines, annotators used their best judgment. Despite the inherent ambiguity in story content annotation, manual inspection of the annotated data indicates good quality, as shown in Table \ref{tab:iou}.



\section*{Ethics Statement}
\shortname{} is constructed with YouTube videos under the Standard YouTube License. For the videos, we release a list of YouTube URL the users can use to download the video from YouTube, as is the standard practice \cite{bain2020condensed,miech2019howto100m, caba2015activitynet}. A fair compensation amount for the annotators was determined by the annotation company based on the difficulty level and time needed to annotate a minute of video in each language. On average, we compensate the annotator 2.44 USD for annotating a minute of video. The exact per-minute compensation amount for each language is shown in Table \ref{tab:price} in the Appendix.

In this paper, we collect user-uploaded videos from YouTube, which are summaries of mostly western movies and TV shows. We recognize that movies and TV shows are fictional in nature, and often prioritize dramatic events over faithful representation of real-life scenarios. In addition, the videos may reflect particular bias of the creators of the movie and TV shows or the creators of the summary videos, as well as bias from particular cultures or the time periods of production. 

For these reasons, we urge researchers to take caution when attempting to learn social norms from such videos. For example, events of bank robberies may be over-represented in these videos, and a machine learning model may inadvertently infer that robbing a bank is part of the social norm. In addition, the model may incorrectly learn from disproportional association of certain groups of people with certain social status, occupations, and other cultural constructs. The dataset is intended for fundamental research and not real-world deployment.

\bibliography{custom}
\clearpage

\appendix
\begin{figure}[!tp]
    \centering
    \includegraphics[width=0.45\textwidth]{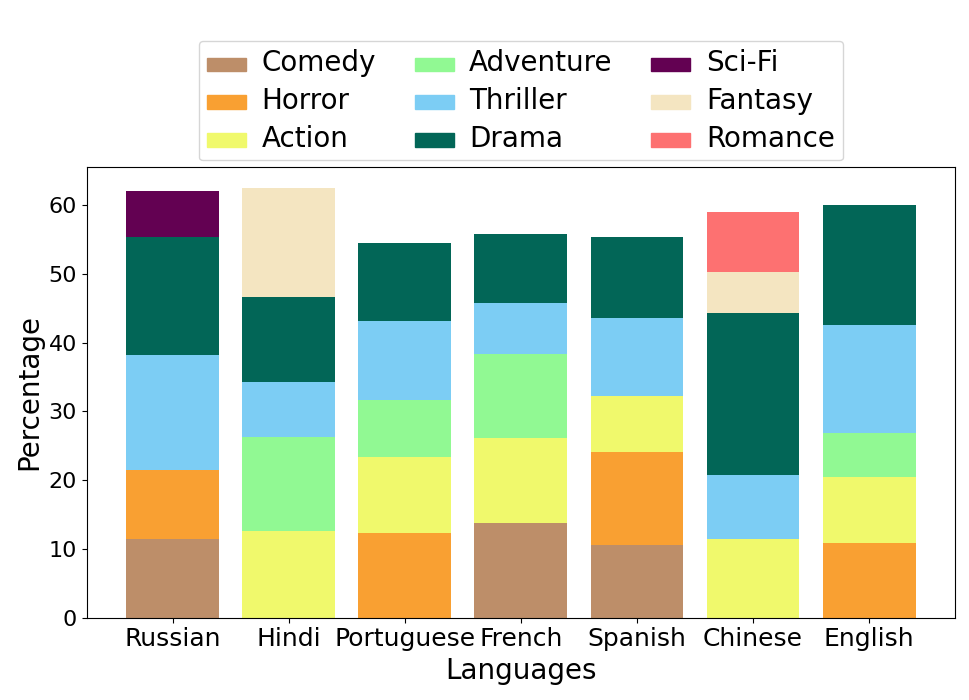}
    \caption{The top 5 genres for movies recapped in each language}
    \label{fig:genre}
\end{figure}

\begin{figure}[t]
    \centering
    \includegraphics[width=0.48\textwidth]{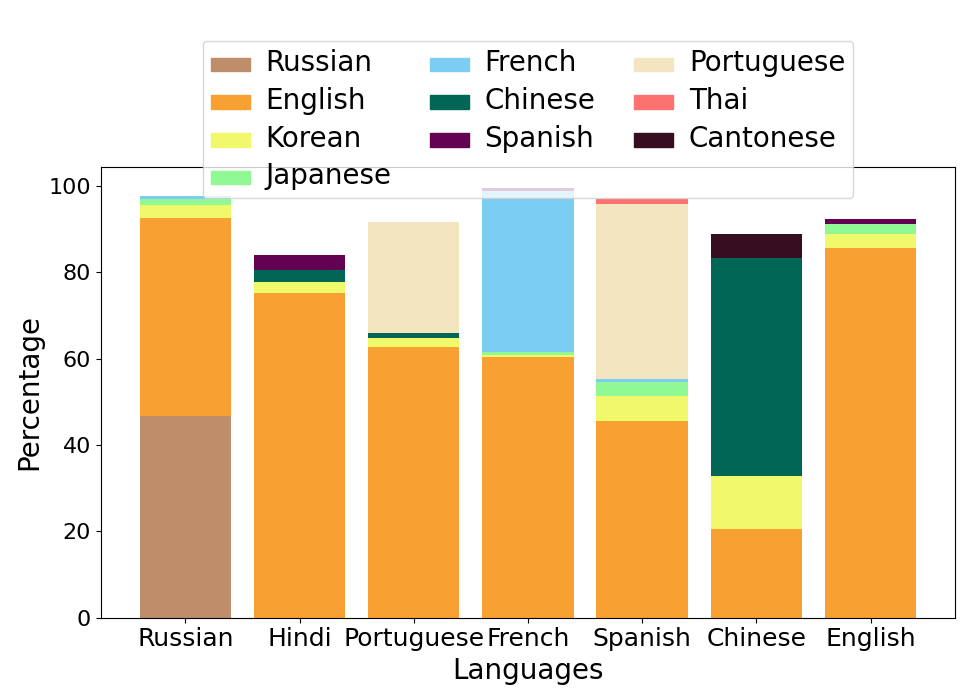}
    \caption{Top 5 languages of release for movies recapped in each language}
    \label{fig:language}
\end{figure}

\begin{figure}[t]
    \centering
    \includegraphics[width=0.48\textwidth]{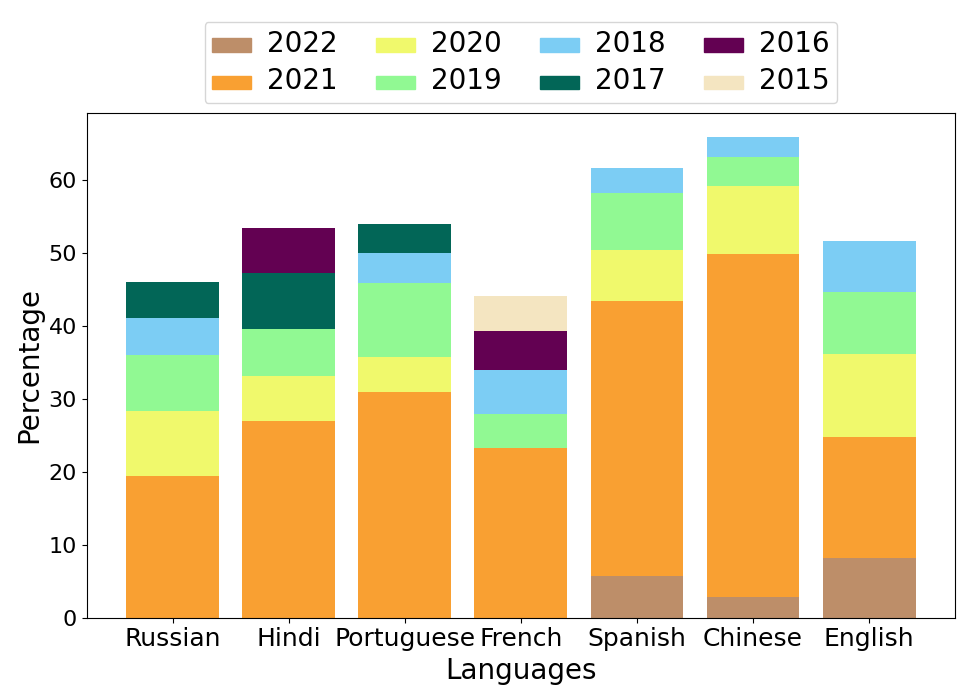}
    \caption{Top 5 years of release for movies recapped in each language}
    \label{fig:year}
\end{figure}

\section{Additional Information on \shortname{}}
\subsection{Additional Preprocessing}
\label{app:remove_overlap}
\paragraph{Removing Overlap Between Train and Test Sets.}
In movie recap videos, videos summarizing the same movie may have similar visual and textual content. 
Thus, we remove all movies that appear in the test sets from the training sets. Specifically, we use ChatGPT to identify the corresponding movie for each video and translate the movie name to English. 
We then remove the overlapped videos based on the English movie names.

\subsection{Additional Analysis}
\label{app:analysis}
As shown in Figure \ref{fig:genre}, creators using different languages prefer similar genre. Although interestingly, ``Fantasy'' films are featured more in Hindi and Chinese than other language, perhaps due to the significant role mythology plays in Eastern cultures.

In Figure \ref{fig:language}, we list the language of release for movies recapped in each language. This trend largely follows the same pattern as the country of release likely because most countries produce movies in their native language.

For each language, we list the top 5 years the movies were released (Figure \ref{fig:year}). As our dataset collection ends around February 2022, \shortname{} largely contains movies released from 2017 to 2021.

\begin{figure}[t]
    \centering
    \includegraphics[width=0.48\textwidth]{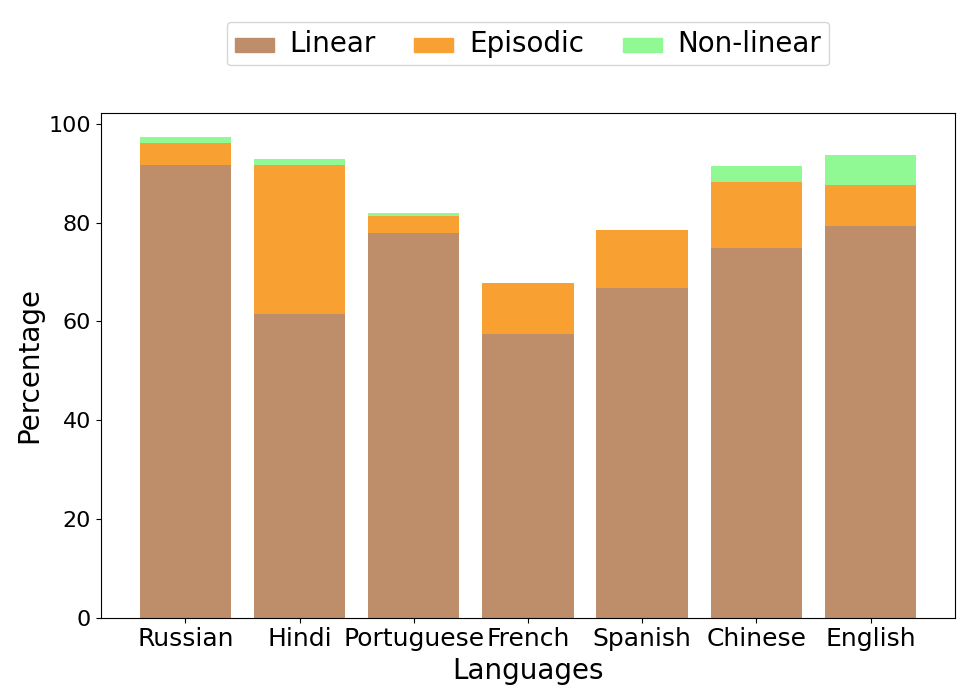}
    \caption{Narrative style of movies recapped in each language}
    \label{fig:narrative}
\end{figure}

In Figure \ref{fig:narrative}, we show the narrative styles of movies recapped in each language. The most common narrative styles are linear and episodic. 

\section{Annotation Details}
\label{app:annotation}
\paragraph{Characteristics Of Annotators.} We employ a team of annotators with professional qualifications in translation for annotation. The team is from Korea. We discussed the compensation amount with the annotation team and set an adequate amount. 

\paragraph{Annotation Instruction.}
We gave the annotators an instruction, shown in Figure \ref{fig:instruction}, and a set of annotation guidelines for ambiguous situations. 
\paragraph{}
\begin{figure}[t]
\begin{tcolorbox}
\small
\tt
\noindent You will see a 5-15 minute video, which is a summary of a movie or a TV show. You will be given a list of sentences that describe the content of the video. Please align each sentence to the video content by writing down the start time and end time of the video segment that matches the sentence. If the sentence does not match any video content, please mark it as ``Unmatched''.
\end{tcolorbox}

\caption{Human annotation instruction.}
\label{fig:instruction}
\end{figure}

\paragraph{Annotation Procedure.}
We divide the data randomly into 4 equal batches, each batch contains a quarter of data from each language. After the annotators annotate each batch, we randomly select 2 videos from each language for validation and employ a student familiar with the task to annotate them. Then, we calculate the IoU between the student's annotations and the annotations provided by the annotators. Specifically, for each sentence we calculate the IoU between the duration annotated by the student and the duration annotated by the annotators. If the IoU on a particular language falls below 60\%, the annotators were asked to redo the annotations for that language in that batch. The average IoU of each language is shown in Table \ref{tab:iou}

\begin{table*}[t]
\centering
\small
\begin{tabular}{@{}p{1.0cm}p{1.0cm}p{1.0cm}p{1.0cm}p{1.0cm}p{1.2cm}p{1.0cm}p{1.0cm}@{}}
\toprule
&English & Chinese & Spanish & French &Portuguese & Hindi & Russian \\
\midrule
IoU &83.2\% & 78.8\% & 84.4\% & 81.3\% & 88.9\% & 84.4\% & 80.8\% \\

\bottomrule
\end{tabular}
\caption{IoU between annotator annotations and our annotations.}
\label{tab:iou}
\end{table*}

\begin{table*}[t]
\centering
\small
\begin{tabular}{@{}p{2.4cm}p{1.0cm}p{1.0cm}p{1.0cm}p{1.0cm}p{1.2cm}p{1.0cm}p{1.0cm}@{}}
\toprule
&English & Chinese & Spanish & French &Portuguese & Hindi & Russian \\
\midrule
Annotation Cost &1.60 & 1.44 & 2.83 & 3.78 & 2.83 & 1.79 & 2.83 \\

\bottomrule
\end{tabular}
\caption{Annotator compensation amount for annotating a minute of video, in USD.}
\label{tab:price}
\end{table*}

\section{Multilingual Punctuation Restoration}
\label{app:punc_restore}
\vspace{0.1in}
\noindent \textbf{Task.}
We treat punctuation restoration as a token classification task, where the model predicts the appropriate punctuation, if any, to follow each token in the sequence. To simplify the task, we limit the predictions to three punctuations: period, comma, and question mark. Other punctuations are either removed (in the case of \#,@," and ') or replaced with periods and commas (in the case of !, ; and :).

\paragraph{Dataset.}
For training, we use the Wiki-40B \cite{guo-etal-2020-wiki} dataset containing clean Wikipedia articles in 40+ languages. For Hindi, we further supplement the training set with data from Wikipedia Hindi \footnote{https://www.tensorflow.org/datasets/catalog/wikipedia}.

To evaluate our punctuation restoration model on narrative content, we build an evaluation set from Wikipedia plot summaries. Specifically, we first identify Wikipedia articles of popular movies. Then, we extract the ``Plot Summary'' portion of the articles. Finally, we remove any overlap between training and test data. We acquire the evaluation sets for French, Spanish, Portuguese, Chinese, and Russian following the above procedure. However, for Hindi, we can not find enough Wikipedia plot summaries. Therefore, we use the wiki-40B validation set for evaluation which may inflate performance on Hindi.

\paragraph{Model.}
Following \cite{frank2021fullstop}, we finetune on the XLM-RoBERTa-large \cite{roberta} model. 

\paragraph{Evaluation.}
We evaluate the model performance with the F1 score.

\paragraph{Baselines.}
We use the Full-Stop \cite{frank2021fullstop} model as the baseline \footnote{The original Full-Stop model is trained on English, German, French, and Italian. We follow their instruction to train on Spanish, French, and Portuguese}. We use the same backbone as Full-Stop, the difference is that they train on the Europarl \cite{koehn2005europarl} dataset while we train on Wiki-40B. The Europarl dataset contains transcripts of political talks in European languages and is commonly used in training multilingual punctuation restoration models.
 
\paragraph{Results.}
Table \ref{tab:punc_result} shows the multilingual punctuation model performance. Our model achieves good performance on all languages. On European languages, our model outperforms Full-Stop by 6-14\% (see Table \ref{tab:punc_result_euro}). This demonstrates that our model is suitable for punctuation restoration in narratives. The performance on Hindi is superior to other languages, this is likely because Hindi is evaluated on in-domain data from Wiki-40B.

\begin{table}[t]
\centering
\begin{tabular}{@{}p{2.0cm}lllll@{}}

\toprule
           & 0    & .    & ,   &?     & Total \\
\midrule
French     & 0.98 & 0.86 & 0.71 & 0.56 & 0.78  \\

Spanish    & 0.97 & 0.77 & 0.67 & 0.53 & 0.74  \\

Portuguese & 0.98 & 0.83 & 0.72 & 0.50 & 0.76  \\

Russian    & 0.96 & 0.73 & 0.77 & 0.51 & 0.74  \\
Chinese    & 0.99 & 0.82 & 0.83 & 0.61 & 0.81  \\
Hindi      & 0.99 & 0.94 & 0.88 & 0.77 & 0.90   \\
Average    & 0.98 & 0.82 & 0.76 & 0.58 & 0.79   \\
\bottomrule
\end{tabular}
\caption{Result of multi-lingual punctuation. }
\label{tab:punc_result}
\end{table}

\begin{table}[t]
\centering
\begin{tabular}{@{}p{2.0cm}lllll@{}}

\toprule
           & 0    & .    & ,   &?     & Total \\
\midrule
\multicolumn{6}{c}{\emph{Wiki-40B (Ours)}}\\
French     & 0.98 & 0.86 & 0.71 & 0.56 & 0.78  \\
Spanish    & 0.97 & 0.77 & 0.67 & 0.53 & 0.74  \\
Portuguese & 0.98 & 0.83 & 0.72 & 0.50 & 0.76  \\
\midrule
\multicolumn{6}{c}{\emph{Europarl}}\\
French &0.97 & 0.80 & 0.64 & 0.47 & 0.72 \\
Spanish & 0.97 & 0.70 &0.59 & 0.15 &0.60 \\
Portuguese & 0.97 &0.75 &0.63 & 0.37 &0.68\\

\bottomrule
\end{tabular}
\caption{Comparison between training on Wiki-40B and training on Europarl. Europarl only contains European languages.}
\label{tab:punc_result_euro}
\end{table}

\begin{table*}[tp]
    \centering
    \footnotesize
    
    \begin{tabular}{@{}p{4.0cm}p{1.0cm}p{1.0cm}p{1.0cm}p{1.0cm}p{1.2cm}p{1.0cm}p{1.0cm}p{1.0cm}@{}}
        \hline
        Model & English & Chinese & Spanish & French & Portuguese & Hindi & Russian & Average \\
        \midrule
        \multicolumn{9}{c}{\textit{Weakly supervised}} \\
        CCLM-multilingual & 0.5 & 7.9 & 1.6 & 1.1 & 1.2 & 1.0 & 1.8 & 2.2 \\
        CCLM-individual & 3.3 & 6.5 & 3.3 & 1.8 & 2.9 & 2.0 & 3.7 & 3.4 \\
        CCLM-translate & 1.7 & 4.7 & 2.4 & 1.9 & 1.9 & 2.1 & 3.1 & 2.5 \\
        CCLM-two-stage & 3.6 & 11.5 & 3.6 & 3.0 & 3.2 & 2.4 & 5.3 & 4.7 \\
        \midrule
        \multicolumn{9}{c}{\textit{Supervised}} \\
        CCLM-translate-supervision & 2.1 & 3.1 & 2.6 & 2.6 & 1.8 & 2.2 & 3.8 & 2.6 \\
        CCLM-two-stage supervision & 3.7 & 13.8 & 3.6 & 3.3 & 3.0 & 2.9 & 4.2 & 4.9 \\
        \hline
    \end{tabular}
    \caption{Text-to-Video retrieval results based on Retrieval@1.}
    \label{tab:t2v-retrieval-1}
\end{table*}

\begin{table*}[t]
    \centering
    \footnotesize
    
    \begin{tabular}{@{}p{4.0cm}p{1.0cm}p{1.0cm}p{1.0cm}p{1.0cm}p{1.2cm}p{1.0cm}p{1.0cm}p{1.0cm}@{}}
        \hline
        Model & English & Chinese & Spanish & French & Portuguese & Hindi & Russian & Average \\
        \midrule
        \multicolumn{9}{c}{\textit{Weakly supervised}} \\
        CCLM-multilingual & 1.6 & 34.5 & 4.9 & 6.2 & 4.3 & 3.9 & 1.6 & 8.1 \\
        CCLM-individual & 10.6 & 17.4 & 10.3 & 5.3 & 7.9 & 6.6 & 10.6 & 9.8 \\
        CCLM-translate & 6.0 & 12.9 & 7.5 & 6.6 & 6.1 & 8.5 & 6.0 & 7.7 \\
        CCLM-two-stage & 10.7 & 30.3 & 12.2 & 8.8 & 10.0 & 8.3 & 10.7 & 13.0 \\
        \midrule
        \multicolumn{9}{c}{\textit{Supervised}} \\
        CCLM-translate-supervision & 7.1 & 12.2 & 8.3 & 7.6 & 6.2 & 7.5 & 7.1 & 8.0 \\
        CCLM-two-stage supervision & 10.7 & 34.2 & 12.0 & 10.1 & 10.6 & 8.3 & 10.7 & 13.8 \\
        \hline
    \end{tabular}
    \caption{Text-to-Video retrieval results based on Retrieval@5.}
    \label{tab:t2v-retrieval-5}
\end{table*}

\begin{table*}[tp]
    \centering
    
    \footnotesize
    \begin{tabular}{@{}p{4.0cm}p{1.0cm}p{1.0cm}p{1.0cm}p{1.0cm}p{1.2cm}p{1.0cm}p{1.0cm}p{1.0cm}@{}}
        \hline
        Model & English & Chinese & Spanish & French & Portuguese & Hindi & Russian & Average \\
        \midrule
        \multicolumn{9}{c}{\textit{Weakly-supervised}} \\
        CCLM-multilingual & 2.5 & 38.4 & 8.2 & 3.6 & 7.1 & 7.1 & 9.5 & 10.9 \\
        CCLM-individual & 16.1 & 27.3 & 16.3 & 9.1 & 12.7 & 9.6 & 20.3 & 15.9 \\
        CCLM-translate & 9.6 & 21.1 & 13.2 & 11.1 & 10.2 & 13.6 & 15.6 & 13.5 \\
        CCLM-two-stage & 16.0 & 40.8 & 19.2 & 14.3 & 15.9 & 12.2 & 21.4 & 20.0 \\
        \midrule
        \multicolumn{9}{c}{\textit{Supervised}} \\
        CCLM-translate-supervision & 12.1 & 19.9 & 13.0 & 12.0 & 10.0 & 13.1 & 16.7 & 13.8 \\
        CCLM-two-stage supervision & 16.6 & 45.1 & 18.3 & 15.7 & 16.5 & 13.9 & 21.9 & 21.1 \\
        \hline
    \end{tabular}
    \caption{Text-to-Video retrieval results based on Retrieval@10.}
    \label{tab:t2v-retrieval-10}
\end{table*}

\begin{table*}[t]
    \centering
    \footnotesize
    
    \begin{tabular}{@{}p{4.0cm}p{1.0cm}p{1.0cm}p{1.0cm}p{1.0cm}p{1.2cm}p{1.0cm}p{1.0cm}p{1.0cm}@{}}
        \hline
        Model & English & Chinese & Spanish & French & Portuguese & Hindi & Russian & Average \\
        \midrule
        \multicolumn{9}{c}{\textit{Weakly-Supervised}} \\
        CCLM-multilingual & 107 & 65 & 422 & 805 & 518 & 409 & 367 & 524 \\
        CCLM-individual & 315 & 74 & 220 & 444 & 311 & 289 & 163 & 259 \\
        CCLM-translate & 525 & 105 & 292 & 402 & 344 & 259 & 176 & 300 \\
        CCLM-two-stage & 333 & 55 & 190 & 337 & 255 & 254 & 136 & 223 \\
        \midrule
        \multicolumn{9}{c}{\textit{Supervised}} \\
        CCLM-translate-supervision & 427 & 108 & 237 & 330 & 334 & 249 & 167 & 265 \\
        CCLM-two-stage supervision & 326 & 47 & 182 & 282 & 254 & 246 & 129 & 209 \\
        \hline
    \end{tabular}
    \caption{Text-to-Video retrieval results based on Mean Rank.}
    \label{tab:t2v-mean-rank}
\end{table*}

\begin{table*}[t]
\centering
\small
\begin{tabular}{@{}p{2.5cm}p{0.5cm}p{0.5cm}|p{0.5cm}p{0.5cm}|p{0.5cm}p{0.5cm}|p{0.5cm}p{0.5cm}|p{0.6cm}p{0.6cm}|p{0.5cm}p{0.5cm}|p{0.5cm}p{0.6cm}@{}}
\toprule
  & \multicolumn{2}{c|}{English}   & \multicolumn{2}{c|}{Chinese}  & \multicolumn{2}{c|}{Spanish} & \multicolumn{2}{c|}{French} & \multicolumn{2}{c|}{Portuguese} & \multicolumn{2}{c|}{Hindi} & \multicolumn{2}{c}{Russian}\\
& Clip.  & Sent.  & Clip.  & Sent. & Clip.  & Sent. & Clip.  & Sent.  & Clip.  & Sent.  & Clip.  & Sent.  & Clip.  & Sent. \\
\midrule
\multicolumn{15}{c}{\textit{Weakly supervised}} \\
CCLM-multi  & 13.4 & 7.3 & 23.5 & 12.8 & 19.5 & 8.9 & 17.6 & 8.8 & 17.4 & 7.2 &11.8 & 3.8 & 15.1 & 7.0 \\
CCLM-individual & 33.2  & 22.3  & 36.6 & 24.6  & 25.7 & 13.6 & 24.2 & 11.2 & 24.8 & 12.4 & 17.7 & 7.3 & 21.9 & 11.2 \\ 
CCLM-translate  & 28.4  & 18.4  & 27.7 & 15.9 & 24.0  & 12.5  &24.3 &13.8 & 21.9 & 10.7 & 18.6 & 8.2 & 20.1& 10.8 \\
CCLM-two-stage  & \textbf{33.7} & \textbf{22.6} &\textbf{43.7} &\textbf{33.3} & \textbf{27.2} &\textbf{15.2} & \textbf{27.8} & \textbf{16.3} & \textbf{27.0} & \textbf{14.6} & \textbf{20.9} & \textbf{9.9}& \textbf{24.5} & \textbf{13.3}\\
\midrule
\multicolumn{15}{c}{\textit{Supervised}} \\
CCLM-translate-s & 30.9 & 20.0 & 27.5 & 20.1 & 26.5 & 14.0 & 26.9 & 15.6 & 24.4 & 12.3 & 20.9 & 9.3 & 21.9 & 11.8 \\
CCLM-two-stage-s & \textbf{34.6} & \textbf{23.5} & \textbf{45.2} & \textbf{34.4} &\textbf{27.4} & \textbf{15.9} & \textbf{28.3} & \textbf{16.9} & \textbf{27.7} & \textbf{15.4} & 20.5 & 9.6 &\textbf{25.1} & \textbf{13.7}\\
\bottomrule
\end{tabular}
\caption{Intra-lingual results based on Clip Accuracy and Sentence IoU. Due to space limitation, we write Clip Accuracy as ``Clip.'', Sentence IoU as ``Sent.'' , ``-supervised'' as ``-s'', and ``multilingual'' as ``multi''.}
\label{tab:iid_results_clip_acc}
\end{table*}
In Table \ref{tab:iid_results_clip_acc}, we show the Clip Accuracy and Sentence IoU for the intra-lingual experiments.

In Tables \ref{tab:clip_acc} and \ref{tab:sent_iou} we show the Clip Accuracy and Sentence IoU scores for cross-lingual transfer, corresponding to Figure \ref{fig:cross-lingual}

\begin{table*}[t]
\centering
\small
\begin{tabular}{@{}p{4.0cm}p{1.0cm}p{1.0cm}p{1.0cm}p{1.0cm}p{1.2cm}p{1.0cm}p{1.0cm}@{}}
\toprule
  & English     &  Chinese & Spanish & French & Portuguese & Hindi & Russian\\
\midrule
CCLM-translated & 28.4   & 27.7  & 24.0    &24.3  & 21.9  & 18.6  & 20.1 \\
\hline
\hline
\rule{-2pt}{2.6ex}
CCLM-two-stage (English) & \textbf{33.7} & 25.4 & 23.3 & 25.9 & 21.1 & 15.1 & 21.4\\
CCLM-two-stage (Chinese) & 22.4 & \textbf{43.7} & 16.7 & 14.8 & 11.8 & 12.0 & 15.3\\
CCLM-two-stage (Spanish) & 32.5 & 29.9 & \textbf{27.2} & 26.6 & 24.2 & 14.8 & 22.0\\
CCLM-two-stage (French) & 31.7 & 30.9 & 25.1 & \textbf{27.8} & 23.1 & 15.0 & 22.5\\
CCLM-two-stage (Portuguese) & 32.6 &30.2 & 25.4 & 27.1 & \textbf{27.0} & 16.6 & 22.2 \\
CCLM-two-stage (Hindi) & 31.8 & 29.8 & 25.2 & 26.1 & 21.7 & \textbf{20.9} & 22.7\\
CCLM-two-stage (Russian) & 33.2 & 31.7 & 25.6 & 26.5 & 23.5 & 16.6 & \textbf{24.5}\\

\bottomrule
\end{tabular}
\caption{Clip Accuracy scores for cross-lingual transfer.}
\label{tab:clip_acc}
\end{table*}

\begin{table*}[t!]
\centering
\small
\begin{tabular}{@{}p{4.0cm}p{1.0cm}p{1.0cm}p{1.0cm}p{1.0cm}p{1.2cm}p{1.0cm}p{1.0cm}@{}}
\toprule
  & English     &  Chinese & Spanish & French & Portuguese & Hindi & Russian\\
\midrule
CCLM-translated   & 18.4   & 15.9   & 12.5   &13.8  & 10.7  & 8.2 & 10.8 \\
\hline
\hline
\rule{-2pt}{2.6ex}
CCLM-two-stage (English) & \textbf{22.6} & 14.2 & 11.7 & 14.8 & 10.8 & 2.9 & 11.5\\
CCLM-two-stage (Chinese) & 13.8 & \textbf{33.3} & 6.7 & 7.6 & 4.2 & 1.4 & 6.6 \\
CCLM-two-stage (Spanish) & 21.5 &18.0 & \textbf{15.2} & 15.6 & 12.8 & 4.6 & 11.7 \\
CCLM-two-stage (French) & 20.8 & 17.8 & 13.1 & \textbf{16.3} & 11.8 & 3.9 & 15.2\\
CCLM-two-stage (Portuguese) & 21.5 & 17.9 & 13.7 & 15.7 & \textbf{14.6} & 5.2 & 11.8 \\
CCLM-two-stage (Hindi)  & 20.8 & 17.4 & 12.5 & 14.8 & 11.0 & \textbf{9.9} & 11.8\\
CCLM-two-stage (Russian) & 22.1 & 19.2 & 13.6 & 15.1 & 12.0 & 5.7 & \textbf{13.3} \\

\bottomrule
\end{tabular}
\caption{Sentence IoU scores for cross-lingual transfer.}
\label{tab:sent_iou}
\end{table*}

\section{Video-Text Retrieval}
\label{app:retrieval}
\vspace{0.1in}
\noindent \textbf{Task.}
Given a set of video clips \( \mathcal{V} = \{ V_1, V_2, \dots, V_M \} \) and a set of textual sentences \( \mathcal{T} = \{ t_1, t_2, \dots, t_N \} \), both from all videos in the test set, the goal is to retrieve the most relevant video or video clips corresponding to each sentence. 

\vspace{0.1in}
\noindent \textbf{Evaluation.}
We evaluate with Retrieval@1 (R@1), Retrieval@5 (R@5), Retrieval@10 (R@5) and Mean Rank.

\vspace{0.1in}
\noindent \textbf{Method.}
For video-text retrieval, we evaluate on the CCLM models trained on \shortname{}. Specifically, we first acquire the video and text embeddings from the video and text encoders. Then we calculate the video text similarity as the cosine similarity between video and text embeddings. Finally, for each text we retrieve the video clip with the highest similarity.

\vspace{0.1in}
\noindent \textbf{Results.}
In Tables \ref{tab:t2v-retrieval-1},\ref{tab:t2v-retrieval-5}.\ref{tab:t2v-retrieval-10},\ref{tab:t2v-mean-rank} we show R@1, R@5, R@10 and MR scores, respectively. 

The text-to-video retrieval results align closely with our alignment results. CCLM-two-stage outperforms CCLM-translate by 2.2-6.5\% on Retrieval @ 1,5,10, demonstrating the advantage of multilingual data. CCLM-two-stage-supervision outperforms CCLM-two-stage by 0.2-1.1\% on Retrieval @ 1,5,10, highlighting the utility of our human annotations. However, supervised finetuning does not improve Spanish performance. We believe this may be due to the greater distribution difference between weakly supervised correspondence, which rely on timestamps, and human-annotated correspondence. Based on our statistics, only 34.0\% timestamp-derived correspondence in Spanish videos match human annotations, while the average across all languages is 44.9\%. Given this large distribution gap, more supervised data may be required to bridge the gap.

\section{Video-Text Alignment}
\subsection{Drop-DTW}
\label{app:drop_dtw}
DTW uses dynamic programming to find the optimal alignment between two sequences based on distance (or similarity), the final alignment corresponds to the shortest distance or highest similarity. In the traditional DTW algorithm, each item in one sequence must match with an item in the other sequence. However, in story videos, some text are not grounded in the video and vice-versa. Therefore, we use the Drop-DTW \cite{dvornik2021drop} algorithm to facilitate dropping certain clips and sentences.

In traditional DTW, to align a sequence of video clips $V=(v_1, \ldots, v_N)$ and a sequence of sentences $T=(t_1, \ldots, t_M)$, we first assume that $v_1$ is aligned to $t_1$. 
Thus, the cost of matching $v_1$ and $t_1$ is $c(1,1)=0$, and the cost of match $v_1$ with $t_j(j \neq 1)$ is $c(1,j)=\infty$, and vice versa. Then, the minimal cost of aligning $(v_1, \ldots, v_i)$ with $(t_1, \ldots, t_j)$, can be calculated as:
\begin{equation}
\begin{split}
    c(i,j) = \min (c(i-1,j)+ d(i,j),\\c(i,j-1)+ d(i,j), \\ c(i-1,j-1)+ d(i,j))  
\end{split}
\end{equation}
where $d(i,j)$ denotes the distance between $v_i$ and $t_j$. Since we have the cosine similarity between each video-text pairs, the distance can be calculated as $d(i,j) = 1 - s(i,j)$, where $s(i,j)$ is the cosine similarity between $v_i$ and $t_j$

In Drop-DTW, we further define two hyperparameters $d_v$ and $d_t$ as the cost of dropping the video and the text respectively. At each time stamp, we calculated the minimal cost as the lowest cost between: (1) matching $v_i$ and $t_j$ and adding $d(i,j)$ to the total cost; (2) dropping $v_i$ and adding $d_v$ to the total cost; (3) dropping $t_j$ and adding $d_t$ to the cost; (4) dropping both $v_i$ and $t_j$ and adding both $d_v$ and $d_t$ to the cost.





\subsection{Addition results}
\label{app:clip_acc}
In Table \ref{tab:iid_results_clip_acc} we show the Clip Accuracy and Sentence IoU scores corresponding to Table \ref{tab:iid_results}.

\subsection{Licensing Information}
The videos we acquire from YouTube are under the Standard YouTube license. Note that in \shortname{}, the videos are released in the form of YouTube URL, and researchers can download the videos directly from YouTube.

For multilingual punctuation restoration we train with data from Wiki-40B \cite{guo-etal-2020-wiki} licensed under the Apache 2.0 License. For scene segmentation, we use TransNet-V2 \cite{souvcek2020transnet} released under the MIT license. ChatGPT, which we use for extracting movie names and metadata, is under the GNU Affero General Public License Version 3.

For video text alignment, the CCLM \cite{zeng2022cross} is under the BSD-3-Clause license and the CLIP4Clip is under the MIT License. YMS \cite{dogan2018neural}, which we use as an out-of-domain evaluation benchmark, is from \url{https://github.com/RubbyJ/Data-efficient-Alignment}.

The usage of all model and data within the paper are in line with their intended uses.

\end{document}